\documentclass[runningheads]{llncs}
\usepackage[T1]{fontenc}
\usepackage{graphicx}
\usepackage{adjustbox}
\usepackage{tabularx}
\usepackage{multirow}
\usepackage{amsmath} 
\usepackage{amssymb} 
\usepackage{times}
\usepackage{latexsym}
\newcommand{\bx}{\mathbf{x}}
\newcommand{\bX}{\mathbf{X}}
\usepackage{booktabs} 
\usepackage{comment}
\usepackage{booktabs}
\usepackage{tabularx}
\usepackage{float}
\usepackage{hyperref}
\usepackage{xcolor}
\definecolor{darkblue}{rgb}{0, 0, 0.5}
 \hypersetup{colorlinks=true, citecolor=darkblue, linkcolor=darkblue, urlcolor=darkblue}
\usepackage{graphicx}  
\usepackage{subcaption} 
\usepackage{algorithm}
\usepackage{algorithmic}
\usepackage[scaled=1.15]{urwchancal}

\usepackage{color} 
\usepackage{ulem} 

\usepackage{graphicx} 
\usepackage{subcaption} 
\usepackage{booktabs} 
\usepackage{float} 

\begin{document}
\title{
Imbalanced Data Clustering via Targeted Data Augmentation Using GMM and LLM
}
\titlerunning{Clustering via Targeted Data Augmentation Using GMM and LLM}
%
\author{Noor Khalal\inst{1},
Abdallah Alaa-Eddine Djamai\inst{2},
Imed Keraghel\inst{1,2}\\
Mohamed Nadif\inst{1}}
\authorrunning{N. Khalal et al.}
%
\institute{Centre Borelli UMR9010, Université Paris Cité, 75006 Paris, France\\ \and
Kernix Software, 75014 Paris, France}
\maketitle       
\begin{abstract}
In Natural Language Processing (NLP), dealing with underrepresented topics is challenging, especially in unsupervised tasks where clustering might not adequately capture minority topics. To tackle this challenge, our paper presents a novel unsupervised data augmentation method that integrates Gaussian Mixture Models (GMMs) and Large Language Models (LLMs). Due to their flexibility and robustness, GMMs can detect clusters corresponding to underrepresented areas in the data, while LLMs create synthetic documents to enrich these clusters and improve their representation. Experiments on various imbalanced text datasets demonstrate that our approach preserves clustering performance in all cases and often enhances cluster interpretability, offering a robust and scalable solution for improving data representation in unsupervised NLP tasks.

\end{abstract}
%
%
%
\section{Introduction}

In unsupervised NLP tasks, the quality of data representation is crucial for effective clustering. However, real-world datasets often contain underrepresented semantic regions, leading to dramatically imbalanced clusters that are challenging for most clustering algorithms. These underrepresented clusters may contain meaningful information that remains poorly captured by traditional clustering algorithms, leading to irrelevant or difficult-to-interpret groupings.

Traditional data augmentation techniques, such as synonym replacement and random insertion \cite{wei2019eda}, aim to increase data diversity by generating new samples from existing ones. Although these methods improve dataset size and variability, they typically apply augmentation uniformly across  the entire dataset, without considering differences in data distribution. This uniformity  does not specifically address sparsely represented regions. Moreover, many existing approaches depend on labeled data to perform class-specific augmentation or resampling \cite{bayer2022survey,karimi2021aeda,guo2019augmenting}, which makes them unsuitable for unsupervised learning scenarios where class labels are unavailable.

Generative models, particularly Large Language Models (LLMs), have recently gained attention for their ability to produce high-quality synthetic text by capturing complex semantic relationships within the data. However, most of the existing work employs LLMs for uniform augmentation across datasets \cite{Ye2024LLMDADA}, which fails to address the imbalance problem.
\begin{figure}
    \centering
    \includegraphics[width=\linewidth]{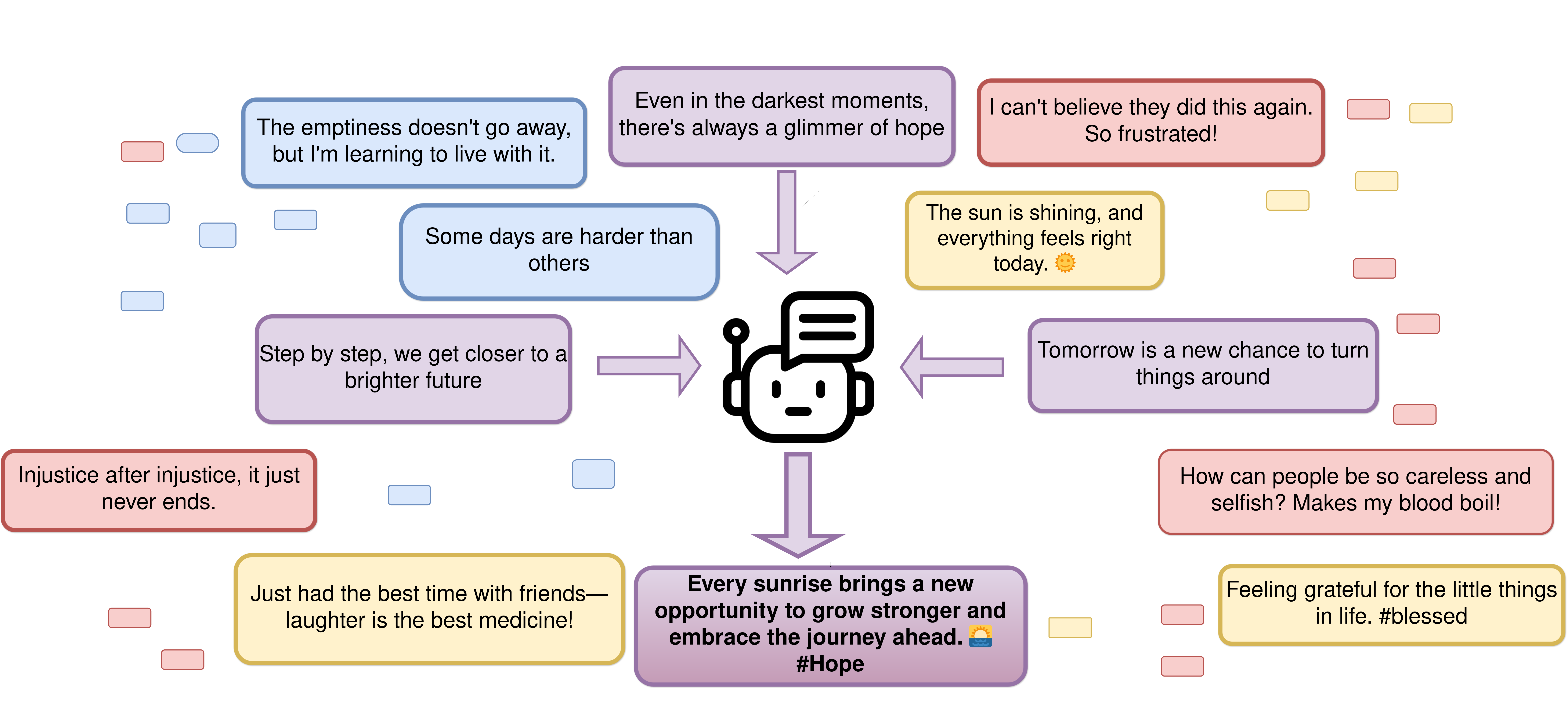}
    \caption{Example from the Tweet Emotion dataset: four emotion clusters (joy in yellow, sadness in blue, anger in red, optimism in purple) where an LLM generates a new tweet about optimism based on three representative samples.}
    \label{fig:tweet_generation}
    
\end{figure}

To overcome these limitations, we present a new method that combines Gaussian Mixture Models (GMMs) \cite{banfield_raftery_1993} and LLMs to perform targeted data augmentation. GMMs are used to analyze the distribution of document embeddings, pinpointing sparse regions that correspond to underrepresented areas in the dataset. LLMs are then used to create synthetic documents in these sparse regions, thereby improving their representation. Figure \ref{fig:tweet_generation} illustrates how, in the \textit{Tweet Emotion} dataset, three tweets about optimism are given to an LLM, which generates a new tweet aligned with the same theme. This process not only adds diversity to the corpus but also enhances its representation. Through comprehensive experiments on various imbalanced text datasets, we show that this augmentation does not diminish clustering algorithm performance. In the worst-case scenario, it maintains the same performance levels, while often leading to enhancements in the interpretability of clusters.

\section{Background and related work}

Data augmentation is a widely used technique in NLP to enhance the diversity and size of training datasets by generating new data from existing samples \cite{qiu2020easyaug,bayer2022survey}. In text classification tasks, traditional augmentation methods such as synonym replacement, random insertion and deletion—collectively known as Easy Data Augmentation (EDA) \cite{wei2019eda}—aim to create varied versions of existing texts to improve model robustness. While these approaches can increase data variability, they often fail to address the issue of class imbalance, as they typically apply augmentation uniformly across all classes.

To tackle class imbalance, techniques such as resampling methods and cost-sensitive learning have been employed \cite{data_generation_llm_imbalanced,karimi2021aeda}. Resampling adjusts the class distribution by replicating minority class samples or reducing majority class samples, while cost-sensitive learning assigns higher misclassification costs to minority classes. However, these methods rely on label information to adjust class distributions or misclassification penalties. In unsupervised settings, where class labels are unavailable, applying these techniques becomes challenging, limiting their effectiveness in balancing the dataset.

Generative models have also been explored for data augmentation in NLP. Generative Adversarial Networks (GANs) \cite{goodfellow2014generative} and their variants, such as Deep Convolutional GANs (DCGANs) \cite{radford2015unsupervised}, have been used to produce synthetic text data by learning the data distribution and generating new samples accordingly. More recently, LLMs have been used to generate diverse textual data for augmentation purposes \cite{kumar2020data,van2023improving}. While these models can produce high-quality text, prior work often employs them to augment data uniformly across all classes without specifically targeting underrepresented ones. For instance, methods like AugGPT \cite{dai2023auggpt} rephrase all documents multiple times but do not address data imbalance, leaving minority classes underrepresented.

Despite these advancements, there remains a gap in the literature regarding data augmentation strategies that combine clustering methods and generative models to target underrepresented topics in unsupervised settings. Existing approaches either require labeled data or do not specifically address class imbalance. Developing effective augmentation strategies for imbalanced data in unsupervised learning remains a challenge and necessitates novel methods that can generate and utilize synthetic data without predefined labels. Our proposal addresses this gap by using GMMs to identify underrepresented clusters and employing LLMs to generate synthetic documents, providing a strategy that requires no labeled data nor prior knowledge of the class structure.

\section{Contribution}

\label{section3}
\begin{figure}[ht]
 \centering

\includegraphics[width=9cm,height=3.9cm]{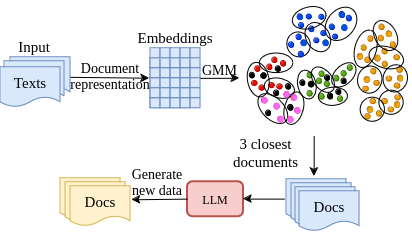}
 \caption{Workflow for generating synthetic documents in a dataset. 1) Text inputs are processed using a Transformer-based embedding model, 2) The EM algorithm derived from GMM is used to cluster these embeddings, 3) Synthetic data points are generated in clusters with the highest volume-to-proportion ratios and their three nearest documents are identified, and 4) The three documents are given to an LLM to generate new textual instances.} 
 \label{fig_description}
\end{figure}

In this section, we outline our approach for augmenting textual data through a combination of embedding techniques and model-based Gaussian methods \cite{banfield_raftery_1993}. The methodology involves several key steps: 1) document representation, 2) clustering with the Expectation-Maximization algorithm (EM) \cite{Dempster_1977} , 3) synthetic data point generation, and 4) using an LLM for data augmentation. The approach is illustrated in Figure~\ref{fig_description}.
\subsection{Document Representation}

To effectively represent documents for clustering, we use a Transformer-based embedding model to generate embeddings that capture rich semantic information. 
These embeddings are particularly suitable for clustering tasks because they enhance the ability to identify subtle semantic connections between unlabeled textual data \cite{imed_embeddings}. 

 We use Uniform Manifold Approximation and Projection (UMAP) \cite{umap2018} for dimensionality reduction, to enhance clustering efficiency. UMAP, a nonlinear technique, preserves data structures while reducing dimensions, maintaining essential relationships and reducing computational complexity. This allows clustering algorithms like GMMs to operate more efficiently by simplifying and speeding up computations.
\subsection{Targeted clustering with GMM}
\paragraph{\textbf{Clustering with Gaussian Mixture Models.}}
In a finite GMM, the data $\bx_1, \ldots, \bx_n$ are assumed to be a sample of $n$ independent instances of a random variable $\bX$ in $\mathbb{R}^d$ where $d$ is the dimensionality of the space. The density of the data can be expressed as:
$$f(\bx_i;\Theta)=\sum_{k=1}^g \pi_k \varphi_k(\bx_i|\mu_k, \Sigma_k),  \quad \forall i \in \{1, \ldots, n\}$$
where 
$\Theta=(\pi_1, \ldots, \pi_g, \mu_1, \ldots, \mu_g, \Sigma_1, \ldots, \Sigma_g)$, $\varphi_k(\bx_i|\mu_k, \Sigma_k)$ is the $k$th component density for observation $\bx_i$ with parameters $(\mu_k, \Sigma_k)$,  $(\pi_1, \ldots, \pi_{g})$ are the mixing weights or probabilities (such that $\pi_k > 0, \sum_k \pi_k=1$) and $g$ is the number of mixture components. This means that the clusters are ellipsoidal, with the mean vector $\mu_k$ at the center, and other geometric features such as volume, shape, and orientation are determined by the covariance matrix $\Sigma_k$ \cite{banfield_raftery_1993}. To estimate the parameter $\Theta$, the log-likelihood is maximized, which is given by:
\begin{equation*}
  L(\bX;\Theta)=\sum_{i=1}^n \log \left( \sum_{k=1}^g \pi_k \varphi_k(\bx_i|\mu_k, \Sigma_k)\right).
\end{equation*}
Maximization uses the Expectation-Maximization (EM) algorithm, which iteratively maximizes the conditional expectation of the complete log-likelihood given $\Theta'$.
\begin{equation*}
  Q(\Theta|\Theta')=\sum_i\sum_k s_{ik}\log\Big(\pi_k \varphi_k(\bx_i|\mu_k, \Sigma_k)\Big).
\end{equation*}
where $s_{ik}=\frac{\pi_k \varphi_\ell(\bx_i|\mu_k, \Sigma_k)}{\sum_\ell \pi_\ell \varphi_\ell(\bx_i|\mu_\ell, \Sigma_\ell)}$ are the posterior probabilities. 

The GMMs provide valuable insights into clusters. An eigen-decomposition of the covariance matrices \(\Sigma_k = \lambda_k D_k A_k D_k^\top\), where \(\lambda_k\) controls the volume, \(A_k\) specifies the shape with $\det(A_k)=1$, and \(D_k\) determines the orientation. This decomposition allows for the detection of clusters with different shapes, volumes, and orientations. The parameter $\pi_k$ represents the proportion of each cluster.

\begin{figure}[!h]
 \centering
 \includegraphics[width=8cm,height=3.7cm]{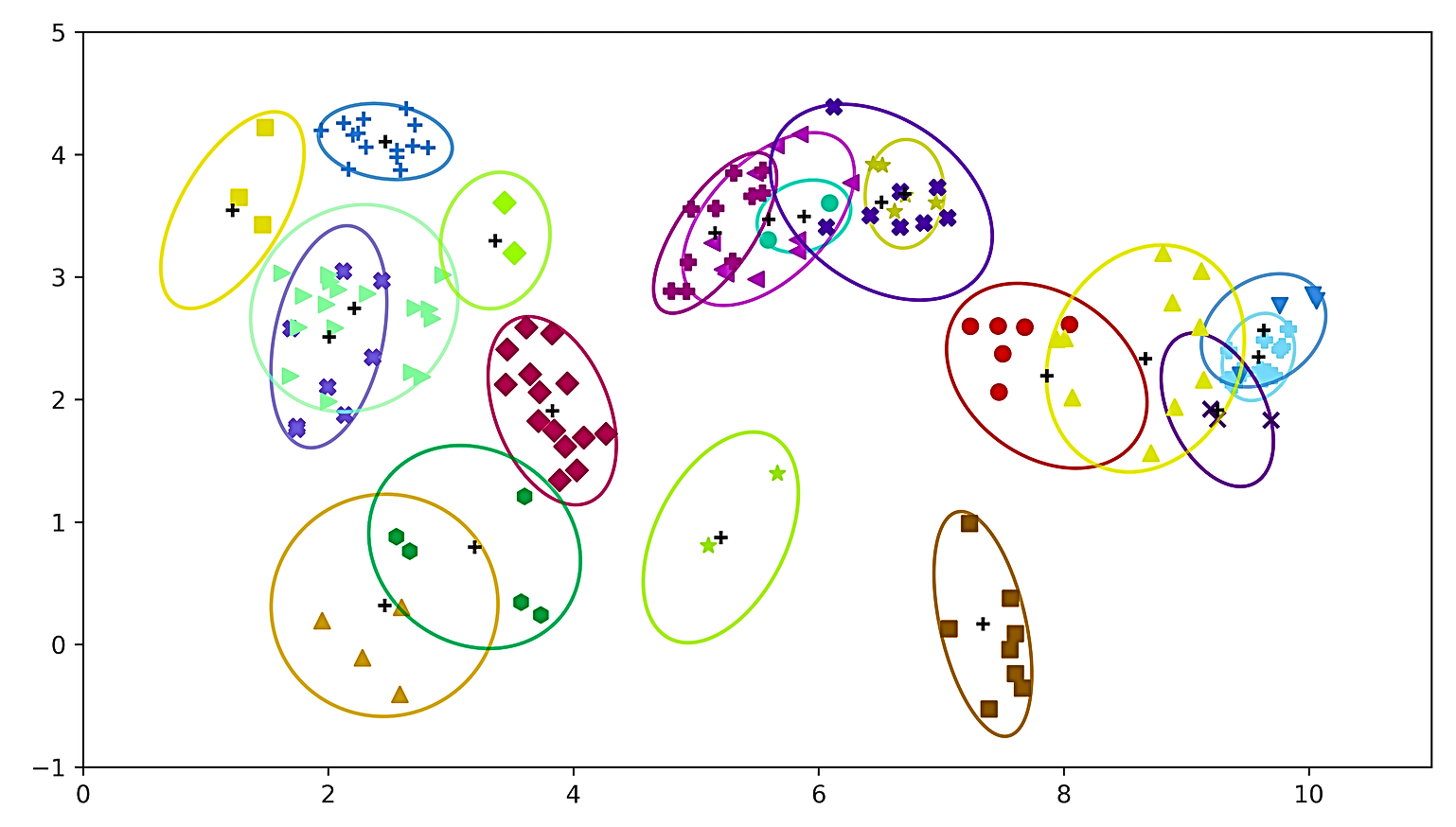}
\caption{Clusters with various orientations, shapes, proportions, and volumes.}
 \label{fig_clusters}
\end{figure}

However, GMMs can face challenges with irregular classes. By increasing the number of clusters, it is possible to capture varying shapes, orientations, sizes, and proportions, as illustrated in Figure~\ref{fig_clusters}, thereby enabling more accurate data modeling. In contrast, K-means optimizes a criterion that assumes spherical classes with equal volumes and proportions. In our work, we fit a GMM to the reduced embeddings. This approach ensures that each cluster contains enough data points to accurately calculate its volume, which is essential for our subsequent analysis.
%
\paragraph{\textbf{Volume and Proportion Analysis.}}
The clusters obtained by EM can be characterized by their proportion and volume. The weight of a cluster, $\pi_k$, represents the proportion of the dataset expected to belong to it, while the volume $V_{k}$ is given by: $$V_{k} = \frac{\pi_k^{d/2}}{d \Gamma(d/2)} \sqrt{\det(\Sigma_k)}$$ where $d$ is the dimensionality of the data, $\Gamma$ is the Gamma function and $\det(\Sigma_k)$ is the determinant of the covariance matrix.

However, when working with high-dimensional low-variance text embeddings such as in our case, the task of calculating the determinant of the covariance matrix can become ill-posed, due to the eigenvalues being close to zero, making the matrix nearly or fully singular and nullifying the determinant, and thus making the computation of cluster volumes impossible. 
\\
To stabilize the calculation and ensure a well-posed problem, Tikhonov regularization \cite{tikhonov1977solutions} is applied by adding a small positive constant to the diagonal elements of the covariance matrix \cite{10.3389/fams.2019.00067}. Mathematically, the regularized matrix is given by $\Sigma_{k}^{'} = \Sigma_k + \alpha I$, where $I$ is the identity matrix and $\alpha$ is the regularization parameter. This adjustment has the effect of shifting the eigenvalues away from zero, thus making the matrix invertible and the determinant (and consequently cluster volume) calculation stable.

\paragraph{\textbf{Identifying underrepresented clusters.}}
Clusters with a high volume but a low proportion highlight regions in the feature space where data points are sparse and spread out, indicating potential underrepresentation. By focusing on clusters with high volume-to-proportion ratios, we target regions where there is significant variability but relatively few data points. This is crucial for targeted data augmentation, where increasing the representation of underrepresented regions can lead to a more balanced and robust dataset.

In our work, we rank all clusters based on their volume-to-proportion ratios to identify underrepresented regions in the data. This approach highlights clusters that have a low proportion of data points (less dense) but occupy a large space in the embedding dimension (higher volumes). By focusing our augmentation on the top half of clusters with the highest volume-to-proportion ratios, we target areas where data is sparse yet potentially significant. This ensures that our data augmentation efforts are directed towards the most \textit{critical} and \textit{underrepresented} regions of the dataset.

\subsection{Targeted Data Augmentation}

\paragraph{\textbf{Synthetic data point generation.}} 
In each \textit{underrepresented cluster} selected as described above, new data points are randomly generated using the Gaussian distribution parameters of the cluster. This approach ensures that the generated points follow the same distribution characteristics as the original data within each cluster. First, the number of samples from each Gaussian component is determined using the multinomial distribution based on the weights $\pi_k$. More specifically,  the total number of samples \( n_{\text{samp}} \) is allocated among the \( g \) components using the multinomial distribution:
$$ P(n_1, n_2, \ldots, n_g) = \frac{n_{\text{samp}}!}{n_1! n_2! \cdots n_g!} \pi_1^{n_1} \pi_2^{n_2} \cdots \pi_g^{n_g}.$$

This results in a set of counts \( (n_1, n_2, \ldots, n_g) \) where each \( n_k \) represents the number of samples to draw from the $k$-th component. 
Then, for each component $k$, samples $X_k$ are drawn from a multivariate normal distribution with mean $\mu_k$ and covariance $\Sigma_k$. This method ensures that the synthetic data points maintain the integrity and structure of each cluster.

\paragraph{\textbf{Textual document generation.}} 
For each sampled point in the reduced representation space, we identify the three closest neighbors among the original documents based on Euclidean distance. These neighbors correspond to real textual documents that are most similar to the sampled point in the embedding space. Using the content of these three nearest documents, we prompt the LLM to generate a new textual document that reflects their combined themes and content. This approach ensures that the generated document is coherent and consistent with the semantic context of the original documents. 

All the steps of our method are outlined in Algorithm~\ref{alg:document_clustering_and_generation}. It is important to note that simply increasing the data size is not always advantageous. Instead, our approach focuses on augmenting clusters that are identified as sparse by GMMs. This ensures that the added data enhances the representation of these clusters rather than merely increasing the overall dataset size.
\begin{algorithm}
\scriptsize
\caption{Document Clustering and Generation}
\label{alg:document_clustering_and_generation} 
\begin{algorithmic}[1]
  \STATE \textbf{Input:} $D=\{d_1, \dots, d_n\} $, $g$ (number of components), $k^*$ number of augmented clusters, $\mathcal{M} $ (an embedding model), $\mathcal{M^*} $ (an instruction-tuned model)
  \STATE $ X_{\text{emb}} \leftarrow \mathcal{M}(D) $ 
  \STATE $ X_R \leftarrow \text{UMAP}(X_{\text{emb}}) $
  \STATE $ \{C_1, \dots, C_g\} \leftarrow \text{EM}(X_R) $
  
  \FOR{$ k \leftarrow 1, g $} 
    \STATE $ S_k \leftarrow \frac{V_{k}}{\pi_{k}} $
  \ENDFOR
  
  \STATE Sort $ \{C_1, \dots, C_g\} $ according to $ S_k $
  \STATE Retain $ k^* $ clusters
  
  \FOR{$ k \leftarrow 1, k^* $}
    \STATE Generate $ (DA)_k $
  \ENDFOR

   \FOR{$ k = 1 \leftarrow k^* $}
    \STATE $ (DA)_k^* = \{\}$
    \FOR{each $ i \in (DA)_k $}
      \STATE Select 3 closest points of $i$ 
      \STATE $ New\_Doc \leftarrow \mathcal{M}^{*}$(3 closest documents)
      \STATE $ (DA)_k^* \leftarrow (DA)_k^* \cup \{ New\_Doc\}$
    \ENDFOR
  \ENDFOR
  \STATE $ (DA)^* \leftarrow \bigcup_{k=1}^{k^{*}} (DA)^{*}_k$

   \STATE $ X_{\text{emb}}^{*} \leftarrow X_{\text{emb}} + \mathcal{M}((DA)^*) $
  \STATE $ X_R^* \leftarrow \text{UMAP}(X_{\text{emb}}^{*}) $
  \STATE $ \{C_1^*, \dots, C_p^*\} \leftarrow \text{Clustering}(X_R^*) $ // such as Kmeans Clustering For Evaluation purposes
\end{algorithmic}

\end{algorithm}

\section{Experiments and results}
Since our approach operates in an unsupervised setting, we evaluate the effectiveness of our targeted data augmentation by applying clustering algorithms to the data. Specifically, we compare the performance of clustering on the original dataset with that on the augmented dataset using baseline methods such as KMean. In this section, we describe our experimental setup and present the results and analyses to demonstrate the effectiveness of our approach.

\subsection{Datasets}

We use five datasets in our experiments, four of which come from the Massive Text Embedding Benchmark (MTEB) \cite{mteb}. 
Each dataset is represented as $ D= (nClasses, \linebreak nDocs, Balance, Avg)$  where $nClasses$ represents the number of groundtruth classes of the dataset, $nDocs$ is the number of total documents of the dataset, $Balance$ is the ratio of the number of documents in the minority class to that in the majority class, expressed in the power of 10, it reflects the degree of class imbalance in the dataset, and $Avg$ is the average number of tokens per document in the datasets.\\
\textbf{Arxiv $(12, 7000, 6.8 \times 10^{-3}, 10)$, Biorxiv $(26, 53787, 4.1 \times 10^{-4}, 13)$,} and \textbf{Medrxiv $(51, 17647,  4.9 \times 10^{-4}, 16)$} are sourced from MTEB \cite{mteb}, specifically the ArxivClusteringS2S, BiorxivClusteringS2S, and MedrxivClusteringS2S datasets\footnote{\url{https://github.com/embeddings-benchmark/mteb/}}. Each dataset consists of article titles from their respective repositories. The cluster labels correspond to categories assigned to the papers by humans. 
\textbf{Reddit $(15, 5114, 1.2 \times 10^{-3}, 11)$}
comprises titles of Reddit posts, sourced from the RedditTitleBody dataset\footnote{\url{https://huggingface.co/datasets/sentence-transformers/reddit-title-body}}. Each post is associated with a subreddit, which serves as the cluster label. The dataset reflects a variety of topics and an unbalanced distribution across subreddits. The dataset
\textbf{Tweet Emotion$(4, 3257, 2.1 \times 10^{-1}, 16)$} consists of tweets labeled with one of four emotions: anger, joy, optimism, and sadness.
\subsection{Experimental Setup}
\paragraph{\textbf{Embedding Generation}}

We employ the \textit{NoInstruct-Small-v0} embedding model\footnote{\url{https://huggingface.co/instructor/NoInstruct-Small-v0}}, which generates embeddings of size 384. This model is selected based on its strong performance on the MTEB leaderboard\footnote{\url{https://huggingface.co/spaces/mteb/leaderboard}}, balancing meaningful semantic representation with computational efficiency. The 384-dimensional embeddings are sufficiently large to capture the nuances of the text while being manageable for subsequent processing.

\paragraph{\textbf{Dimensionality Reduction}}

To facilitate efficient clustering, we apply UMAP to reduce the embeddings to 10 dimensions. UMAP effectively preserves the local and global structure of the data in lower dimensions. Reducing to 10 dimensions ensures that, in the GMM clustering step, the number of dimensions does not exceed the number of data points in any cluster, which is important for stable covariance matrix estimation.

\paragraph{\textbf{Clustering Parameters}}

We fit a Gaussian Mixture Model (GMM) to the reduced embeddings $X_R$. The number of components $g$ is set to $\lceil nDocs^{1/3} \rceil$, following the recommendation in \cite{wong1982hybrid}. We use full covariance matrices to allow each cluster to have its own covariance, providing flexibility in modeling the data distribution. Thus, the parameters are estimated using the EM algorithm, and the clusters are determined based on the maximum a posteriori principle.

\paragraph{\textbf{Augmentation Process}}

After obtaining the GMM clusters, we compute the volume-to-proportion ratio $S_k = \frac{V_k}{\pi_k}$ for each cluster, where $V_k$ is the volume of cluster $C_k$ and $\pi_k$ is its mixture weight. Clusters are sorted in descending order based on $S_k$, highlighting clusters that are large in volume but contain relatively few data points, indicative of underrepresented regions in the data. We select the top $k^*$ clusters with the highest $S_k$ values for augmentation. For each selected cluster, we sample additional points within the cluster's distribution to generate synthetic embeddings $(DA)_k$. For each synthetic embedding, we identify the three nearest original embeddings in $X_R$ using Euclidean distance. The corresponding documents of these neighbors are retrieved to provide context for the LLM. We use \textit{Mistral-7B-Instruct-v0.2} \cite{mistral_article} as our instruction-tuned LLM for text generation. This model is capable of generating coherent and contextually relevant text based on the provided prompts. By feeding the LLM with the texts of the three nearest neighbors, we generate new documents that reflect the combined content of these examples. 
The generated documents $(DA)^*$ are then embedded using the same embedding model $\mathcal{M}$ and added to the original embeddings $X_{\text{emb}}$ to form the augmented dataset $X_{\text{emb}}^{*}$. UMAP is applied again to reduce the dimensionality of the augmented embeddings, resulting in $X_R^*$ for clustering evaluation.

To assess the impact of our targeted data augmentation, we applied two baseline clustering algorithms to both the original and augmented datasets. These algorithms are chosen for their effectiveness and simplicity in clustering applications.
\begin{itemize}
\item \textbf{KMeans:} A widely used algorithm that partitions data into $k$ clusters by minimizing the sum of squares within the cluster. We used the implementation from scikit-learn with 'k-means++' initialization for better convergence and clustering performance.
\item \textbf{Spherical KMeans (SKmeans)} is a variant of KMeans designed for \linebreak high-dimensional data, where data points are normalized to lie on the surface of a unit sphere. It uses cosine similarity instead of Euclidean distance to measure the closeness between points and cluster centroids.
\end{itemize}

\subsection{Results and Analysis}
To evaluate the impact of our targeted data augmentation, we examine three aspects:\linebreak (1) cluster distribution, to observe how augmentation influences the balance of clusters; (2) keyword comparison, to assess improvements in thematic coverage and interpretability; and (3) clustering performance, evaluated using two widely used metrics: \textit{Normalized Mutual Information} (NMI) \cite{strehl2002cluster}, which measures the quality of the clustering against ground truth, and \textit{Adjusted Rand Index} (ARI) \cite{steinley2004properties}, which evaluates the similarity between predicted and true cluster assignments while accounting for chance.

\subsubsection{Cluster Distribution}
\begin{figure}[!h]
  \centering
  \begin{subfigure}[b]{0.48\textwidth}
    \centering
    \vspace{-1cm}
\includegraphics[width=\textwidth, height=3.8cm]{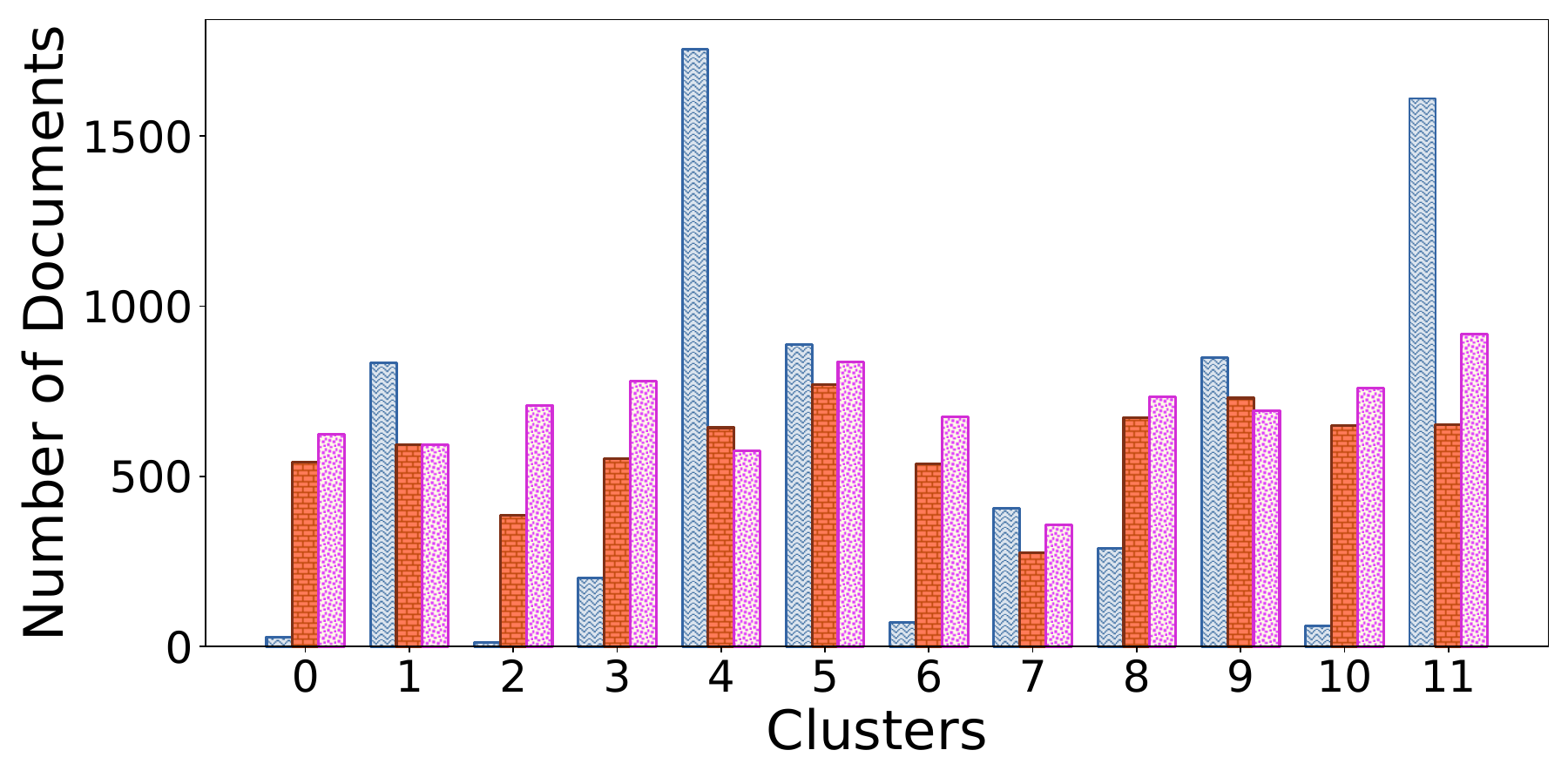}
    \label{fig:image2}
  \end{subfigure}
  \hfill
  \begin{subfigure}[b]{0.48\textwidth}
    \centering
    \includegraphics[width=\textwidth, height=3.8cm]{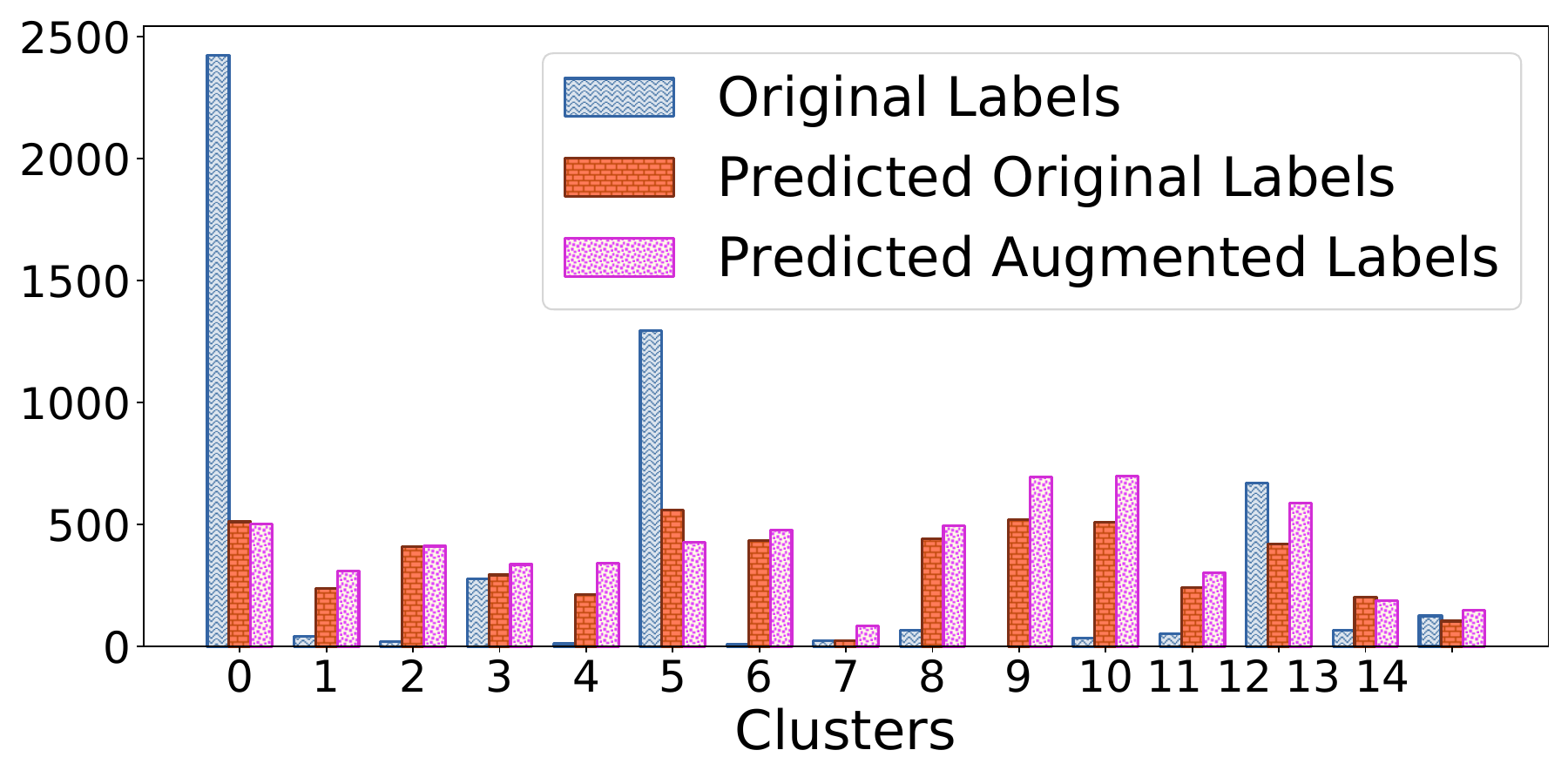}
    \label{fig:image3}
  \end{subfigure}
  \vspace{-0.6cm}
  \caption{ 
    Document distributions across clusters for the Arxiv (left) and Reddit (right) datasets. 
  }
  \label{fig:cluster_distribution}
\end{figure}

Figure \ref{fig:cluster_distribution} illustrates the document distributions across clusters for two datasets: Reddit and Arxiv. These datasets were selected due to their manageable number of clusters (12 and 15, respectively), which allow for a clear visualization of distribution patterns. 
The bar graphs depict three scenarios: the original class labels (blue zigzags), the baseline KMeans clustering applied to the original non-augmented data (red bricks), and the KMeans clustering after data augmentation (pink dots).

In the original labels (blue zigzags), the imbalance of the datasets is evident, with some classes overrepresented and others underrepresented. The baseline KMeans (red bricks) uniformly distributes documents, reducing the dominance of overrepresented classes but not accurately reflecting the initial imbalance.
After data augmentation (pink dots), KMeans continues to equalize cluster sizes. Yet, an important trend emerges: peaks in the augmented data (pink dots) often align with clusters corresponding to minority classes in the original distribution (blue zigzags). 

In summary, data augmentation improves the representation of minority classes, demonstrating the effectiveness of this strategy in addressing data imbalance. However, it is now reasonable to ask two questions: a) What does our methodology contribute in terms of the interpretability of clusters? b) Does this augmentation challenge the quality of the clustering?

\subsubsection{Keyword comparison}
To address the interpretability aspect of the classes, we first rely on the analysis of distribution of keywords in the KMeans clusters on augmented and non-augmented data. The keywords are extracted using \textbf{KeyBERT}; a keyword extraction technique that uses contextual embeddings to identify the most representative terms within a cluster.

\begin{table}[!h]
\centering
\scriptsize
\caption{\footnotesize Comparison of top words and their frequencies in overrepresented (+) and underrepresented (-) clusters before and after data augmentation. The number at the bottom right of each word represents the frequency of the keyword in the dataset vocabulary. Words in bold are new terms introduced in the augmented data, while the rest of the words are shared between both datasets and are underlined when they have the higher frequency. }
\begin{adjustbox}{width=\textwidth}
\begin{tabular}{|c|p{3.15cm}|p{3.15cm}|p{4.2cm}|}
\hline
\textbf{Cluster} & {\phantom{ texttt} \textbf{Non-Augmented}} & \textbf{\phantom{ textttttt}Augmented} & \textbf{\phantom{ texttttttttt}Changes Observed} \\ \hline
\multirow{5}{*}{\textbf{\rotatebox{90}{Optimism (-)}} } & 
\begingroup\spaceskip=0.1em 
depression$_{58}$,
life$_{38}$,
day$_{26}$, 
feel$_{24}$,
nervous$_{22}$, 
lost$_{21}$,
panic$_{16}$,
optimism$_{16}$, 
despair$_{15}$, 
gloomy$_{15}$, 
love$_{14}$, 
shy$_{13}$ \endgroup & 

\begingroup\spaceskip=0.1em
depression$_{58}$, 
{{\textbf{music}}$_{40}$}, 
\uline{life}$_{40}$,
\uline{day}$_{36}$, 
\uline{feel}$_{25}$, 
\uline{lost}$_{22}$, 
{{\textbf{happy}}$_{22}$}, 
\uline{love}$_{21}$, 
\uline{optimism}$_{17}$,
{{\textbf{sober}}$_{17}$}, 
{{\textbf{smile}}$_{16}$}, 
{{\textbf{birthday}}$_{15}$} \endgroup & 

\begingroup\spaceskip=0.2em 
Augmentation expanded the focus to more optimistic and positive themes, introducing diversity while reducing emphasis on negative or neutral terms.
\endgroup
\\ \hline

\multirow{5}{*}{\textbf{\rotatebox{90}{Anger (+)}} } & 
\begingroup\spaceskip=0.1em 
angry$_{41}$,
bully$_{35}$,
\uline{outrage}$_{30}$,
{terror}$_{28}$,
{people}$_{26}$, 
rage$_{25}$, 
dont$_{24}$, 
offended$_{23}$,
revenge$_{17}$, 
irritate$_{17}$, 
insult$_{17}$,
hate$_{13}$ \endgroup & 

\begingroup\spaceskip=0.001em 
angry$_{41}$,
bully$_{35}$,
\uline{terror}$_{29}$, 
\uline{people}$_{28}$, 
{outrage}$_{27}$, 
rage$_{25}$, 
insult$_{17}$, 
hate$_{13}$, 
{{\textbf{game}}$_{10}$}, 
{{\textbf{play}}$_{10}$}, 
revenge$_{9}$, 
{{\textbf{racism}}$_{9}$} \endgroup & 

\begingroup\spaceskip=0.2em 
Terms didn’t change significantly, but some new ones were added, expressing more intense anger (e.g., \textit{game}, \textit{racism}). Overall, the cluster remained largely consistent. \endgroup 
\\ \hline

\multirow{5}{*}{\textbf{\rotatebox{90}{Skincare (-)}} } & 
\begingroup\spaceskip=0.1em 
sellus$_{6}$,
farmacy$_{2}$,
moisturizer$_{1}$,
brand$_{1}$,
moisture$_{1}$,
birthday$_{1}$,
glossier$_{1}$,
balm$_{1}$,
rituals$_{1}$,
babor$_{1}$,
small$_{1}$,
look$_{1}$,
 \endgroup & 

\begingroup\spaceskip=0.1em 
\uline{sellus}$_{8}$
{\textbf{skincare}$_{6}$},
\uline{sale}$_{5}$,
\uline{glossier}$_{4}$,
{\textbf{cream}$_{4}$},
{{\textbf{toner}$_{4}$}},
\uline{balm}$_{3}$,
{\textbf{treatments}$_{3}$},
{\textbf{dark}$_{3}$},
{\textbf{spot}$_{3}$},
{\textbf{sun}$_{3}$},
farmacy$_{2}$,
\uline{moisturizer}$_{2}$,
\endgroup & 

\begingroup\spaceskip=0.2em 
Augmentation added skincare-specific terms (e.g., \textit{skincare}, \textit{cream}, \textit{toner}) and shifted focus slightly toward sales (\textit{sale}). \endgroup 
\\ \hline

\multirow{5}{*}{\textbf{\rotatebox{90}{DogeCoin (+)}} } & 
\begingroup\spaceskip=0.1em 
doge$_{250}$,
bought$_{17}$,
currency$_{11}$,
market$_{7}$,
coin$_{7}$,
community$_{7}$,
dollar$_{6}$
 \endgroup & 

\begingroup\spaceskip=0.1em 
\uline{doge}$_{254}$,
\uline{bought}$_{27}$,
\uline{currency}$_{17}$,
\uline{dollar}$_{11}$,
\uline{coin}$_{11}$,
\uline{community}$_{11}$,
price$_{10}$,
\uline{market}$_{8}$
\endgroup & 

\begingroup\spaceskip=0.2em 
Terms remained consistent, with minor increases in frequencies. Augmentation slightly emphasized financial aspects, but the dominant focus on doge was unchanged. \endgroup 
\\ \hline

\multirow{6}{*}{\textbf{\rotatebox{90}{Economics (-)}} } & 
\begingroup\spaceskip=0.00001em 
sustainable$_{5}$,
inequality$_{4}$,
frequency$_{4}$,
likelihood$_{3}$,
dimensional$_{2}$,regression$_{2}$
pandemic$_{1}$,governments$_{1}$
evolutionary$_{1}$,strategy$_{1}$,
approach$_{1}$,macroecon$_{1}$
 \endgroup & 

\begingroup\spaceskip=0.01em 
\textbf{model}$_{7}$,
\uline{regression}$_{5}$,
\textbf{series}$_{4}$,
\uline{dimensional}$_{4}$,
\textbf{learning}$_{3}$,
\uline{pandemic}$_{3}$,
\textbf{sparsity}$_{2}$,
\uline{sustainable}$_{2}$,
\uline{approach}$_{2}$,
\uline{likelihood}$_{1}$
\endgroup & 

\begingroup\spaceskip=0.2em 
The augmentation introduced new terms (e.g., \textit{series}, \textit{sparsity}) reflecting broader methodological aspects of economics. Existing terms saw increased emphasis, indicating a stronger focus on statistical applications. 
 \endgroup 
\\ \hline

\multirow{5}{*}{\textbf{\rotatebox{90}{CS (+)}} } & 
\begingroup\spaceskip=0.1em 
learning$_{47}$,
neural$_{21}$,
networks$_{20}$,
classification$_{17}$,
adversarial$_{15}$,
deep$_{15}$,
models$_{14}$,
detection$_{12}$,
recognition$_{11}$
 \endgroup & 

\begingroup\spaceskip=0.1em 
\uline{learning}$_{74}$,
\uline{neural}$_{43}$,
\uline{detection}$_{35}$,
\uline{networks}$_{31}$,
\uline{deep}$_{31}$,
\uline{classification}$_{30}$,
\uline{recognition}$_{27}$,
\uline{adversarial}$_{22}$,
\uline{models}$_{21}$,
\endgroup & 

\begingroup\spaceskip=0.2em 
The words remain consistent, with increased emphasis on core computer science terms like \textit{learning}, \textit{neural}, and \textit{detection}.
 \endgroup 
\\ \hline

\multirow{6}{*}{\textbf{\rotatebox{90}{A B\&C (-)}} } & 
\begingroup\spaceskip=0.1em 
drosophila$_{14}$,
olfactory$_{10}$,
neurons$_{6}$,
circadian$_{4}$,
cortex$_{4}$,
pathway$_{3}$,
learning$_{3}$,
dopaminergic$_{3}$,
model$_{3}$,
light$_{2}$,
endocrine$_{2}$
 \endgroup & 

\begingroup\spaceskip=0.1em 
\uline{drosophila}$_{257}$,
\uline{olfactory}$_{176}$,
\uline{circadian}$_{166}$,
\textbf{sleep}$_{66}$,
\textbf{clock}$_{62}$,
\uline{neurons}$_{54}$,
\textbf{taste}$_{47}$,
\uline{light}$_{30}$,
\textbf{sensory}$_{26}$,
\textbf{rhythms}$_{24}$,
\textbf{odorant}$_{24}$
\endgroup & 
New terms like \textit{sleep}, \textit{clock}, and \textit{taste} introduce broader behavioral contexts. Existing words such as \textit{drosophila} and \textit{olfactory} gained prominence, emphasizing sensory and neural mechanisms in animal cognition.
\begingroup\spaceskip=0.2em 
 \endgroup 
\\ \hline

\multirow{6}{*}{\textbf{\rotatebox{90}{Neuroscience{\tiny(+)}}} } & 
\begingroup\spaceskip=0.1em 
neural$_{69}$,
cortex$_{58}$,
learning$_{39}$,
visual$_{38}$,
memory$_{30}$,
auditory$_{25}$,
temporal$_{19}$,
cortical$_{19}$,
speech$_{17}$,
spatial$_{16}$,
dynamics$_{15}$,
prefrontal$_{15}$
 \endgroup & 

\begingroup\spaceskip=0.1em 
\uline{neural}$_{546}$,
\uline{cortex}$_{407}$,
\uline{visual}$_{318}$,
\uline{learning}$_{280}$,
\textbf{brain}$_{251}$,
\uline{memory}$_{244}$,
\uline{cortical}$_{195}$,
\uline{auditory}$_{168}$,
\textbf{attention}$_{132}$,
\textbf{perception}$_{121}$,
\uline{dynamics}$_{119}$,
\uline{speech}$_{117}$
\endgroup & 

\begingroup\spaceskip=0.2em 
The cluster shows a substantial increase in emphasis on core terms such as \textit{neural} and \textit{cortex}. New terms like \textit{brain}, and \textit{perception} expand the focus to include cognitive and sensory processes, complementing existing terms like \textit{visual} and \textit{auditory}.
 \endgroup 
\\ \hline

\multirow{6}{*}{\textbf{\rotatebox{90}{Neurology (-)}} } & 
\begingroup\spaceskip=0.1em 
alzheimers$_{35}$,
cognitive$_{14}$,
disease$_{12}$,
dementia$_{10}$,
brain$_{10}$,
genetic$_{5}$,
impairment$_{5}$,
biomarkers$_{4}$,
diagnosis$_{4}$,
amyotrophic$_{4}$,
trials$_{4}$,
risk$_{4}$

 \endgroup & 

\begingroup\spaceskip=0.1em 
\uline{alzheimers}$_{156}$,
\textbf{parkinsons}$_{115}$,
\uline{cognitive}$_{115}$,
\uline{disease}$_{107}$,
\uline{brain}$_{97}$,
\uline{dementia}$_{85}$,
\uline{impairment}$_{45}$,
\textbf{epilepsy}$_{44}$,
\uline{genetic}$_{30}$,
\textbf{eeg}$_{29}$,
\textbf{stroke}$_{28}$,
\textbf{cognition}$_{25}$
\endgroup & 

\begingroup\spaceskip=0.2em 
New terms like {\textit{parkinsons}} and {\textit{stroke}} broaden the scope to include more neurological conditions. Existing words such as {\textit{alzheimers}} and {\textit{dementia}} show significant increases, reinforcing the focus on degenerative diseases and brain function.
 \endgroup 
\\ \hline

\multirow{6}{*}{\textbf{\rotatebox{90}{Epidemiology{\tiny(+)}}} } & 
\begingroup\spaceskip=0.1em 
sarscov2$_{89}$,
covid19$_{28}$,
antigen$_{22}$,
testing$_{19}$,
diagnostic$_{16}$,
rtpcr$_{15}$,
saliva$_{14}$,
rna$_{13}$,
detection$_{10}$,
test$_{9}$,
nasopharyngeal$_{9}$,
viral$_{8}$

 \endgroup & 

\begingroup\spaceskip=0.1em 
\uline{sarscov2}$_{329}$,
\uline{antigen}$_{125}$,
\uline{covid19}$_{104}$,
\uline{detection}$_{70}$,
\uline{testing}$_{70}$,
\uline{saliva}$_{67}$,
\uline{diagnostic}$_{54}$,
\uline{rtpcr}$_{44}$,
\uline{test}$_{39}$,
\uline{rna}$_{32}$,
\textbf{tests}$_{31}$,
\textbf{screening}$_{30}$
\endgroup & 

\begingroup\spaceskip=0.2em 
The focus on diagnostics and testing is reinforced with increases in terms like {\textit{sarscov2}}, {\textit{antigen}}, and {\textit{covid19}}. New additions such as {\textit{tests}} and {\textit{screening}} highlight expanded approaches to epidemiological detection.
 \endgroup 
\\ \hline

\end{tabular}
\end{adjustbox}
\label{tab:keyword_tweet_reddit}
\vspace{-0.6cm}
\end{table}
Table \ref{tab:keyword_tweet_reddit} presents the top 12 keywords (based on their frequency) for each cluster from both augmented and non-augmented data, spanning various datasets: \linebreak \textit{Tweet\_Emotions}, \textit{Reddit}, and the scientific corpora (\textit{Arxiv}, \textit{Biorxiv}, and \textit{Medrxiv}).
The correspondence between clusters in the augmented and non-augmented datasets is
identified using the Hungarian algorithm.
The analysis highlights how augmentation affects underrepresented and overrepresented clusters across different domains. In the \textit{Tweet\_Emotions} dataset, the augmentation enriched the \textit{Optimism} cluster with new positive terms such as \textit{happy} and \textit{smile}, while negative words like \textit{nervous} and \textit{panic} were removed. This resulted in a sharper focus on optimistic themes. For the \textit{Anger} cluster, new terms such as \textit{game} and \textit{racism} were introduced, reflecting broader and more intense expressions of anger. However, much of the cluster remained stable, as expected for an overrepresented category.

In the \textit{Reddit} dataset, the underrepresented \textit{Skincare} cluster became more specific with new terms like \textit{cream} and \textit{toner}, while the overrepresented \textit{Dogecoin} cluster showed minimal changes, maintaining its focus on \textit{doge} with slight increases in words such as \textit{bought} and \textit{currency}.

For the scientific corpora, the augmentation had varied impacts. In the \textit{Arxiv} dataset, the \textit{Economics} cluster introduced terms like \textit{regression} and \textit{sparsity}, broadening its methodological scope. The \textit{Computer Science} cluster remained stable with minor additions. In the \textit{Biorxiv} dataset, the \textit{Animal Behavior and Cognition} (\textit{A B\&C}) cluster introduced sensory-related terms like \textit{sleep} and \textit{taste}, reflecting a focus on sensory mechanisms. In the \textit{Neuroscience} cluster, new terms like \textit{brain} emerge, while core words show a tenfold increase (e.g., \textit{cortex}, increasing from 69 to 546). This reflects the proportional increase in generated data relative to the dataset's large size. In the \textit{Medrxiv} dataset, the neurology cluster highlighted terms like \textit{parkinsons} and \textit{stroke}, emphasizing neurological conditions. The \textit{Epidemiology} cluster introduced terms like \textit{sarscov2} and \textit{testing}, reinforcing diagnostics and epidemic detection.

Overall, the results demonstrate that augmentation effectively enriched underrepresented clusters by introducing domain-specific and diverse terms, improving their interpretability. Overrepresented clusters, as expected, showed minimal changes, preserving their structure while occasionally gaining some frequency shifts.
\vspace{-2mm}
\subsubsection{Clustering performance}
In this section, we present the clustering performance analysis summarized in Table \ref{tab:clustering_performance}. The results compare clustering metrics (NMI and ARI) across our five datasets, for two algorithms: KMeans and SKMeans, using non-augmented and augmented data. Metrics, averaged over five runs, show that augmentation often improves performance, particularly in ARI, and maintains comparable results even in worst-case scenarios, ensuring that clustering quality is not compromised.
\begin{table}[!h]
\vspace{-0.3cm}
    \centering
    \caption{Clustering performance results (mean ± standard deviation). Results are reported for both non-augmented (N.A) and augmented (A) datasets.}
    \begin{tabular}{cccccccc}
    \toprule \\
    \textbf{Algo.} & \textbf{Type} & \textbf{Metric} & \textbf{Tweet\_Emo.} & \textbf{Reddit} & \textbf{Arxiv} & \textbf{Medrxiv} & \textbf{Biorxiv} \\
    \midrule
    \multirow{5}{*}{\textbf{\rotatebox{90}{KMeans}}} 
        &  \multirow{2}{*}{\rotatebox{0}{N.A}} 
         &  NMI & 20.32 ± 2.28 & 54.74 ± 1.73 & 44.22 ± 0.91 &  30.07 ± 0.21 &  \textbf{34.03 ± 0.17} \\ 
         & &  ARI & 22.79 ± 6.23 & 25.69 ± 2.52 & 32.09 ± 2.26 & 6.92 ± 0.35  & 20.09 ± 1.79  \\
         \cmidrule{2-8}
         &  \multirow{2}{*}{\rotatebox{0}{A}} 
          &  NMI & \textbf{22.10 ± 3.49} & \textbf{55.08 ± 0.58} & \textbf{44.28 ± 0.57} & \textbf{34.62 ± 0.21} &  30.39 ± 0.21  \\ 
         & &  ARI & \textbf{22.81 ± 5.83} & \textbf{29.41 ± 1.78} & \textbf{34.10 ± 1.64} & \textbf{7.22 ± 0.24}  &  \textbf{20.59 ± 1.49} \\
    \midrule
    \multirow{5}{*}{\textbf{\rotatebox{90}{SKmeans}}} 
        &  \multirow{2}{*}{\rotatebox{0}{N.A}} 
         &  NMI &  20.85 ± 2.80 & \textbf{55.91 ± 0.83} & 43.59 ± 0.48 & 30.03 ± 0.30 & \textbf{34.12 ± 0.16} \\ 
         & &  ARI & 19.39 ± 6.72 & 29.88 ± 1.83 & 32.75 ± 1.49 & 7.17 ± 0.44 &  22.24 ± 2.25\\ 
         \cmidrule{2-8}
         &  \multirow{2}{*}{\rotatebox{0}{A}} 
         &  NMI & \textbf{22.42 ± 5.04} & 55.46 ± 1.21 & \textbf{44.42 ± 0.72} & \textbf{30.44 ± 0.29} & 33.86 ± 0.29 \\ 
         & &  ARI & \textbf{23.67 ± 9.74} & \textbf{31.53 ± 2.78} & \textbf{33.58 ± 1.60} & \textbf{7.65 ± 0.34} & \textbf{23.36 ± 3.01}\\ 

    \midrule

   \end{tabular}
    \label{tab:clustering_performance}
\end{table}
\section{Conclusion}
In this work, we presented a novel data augmentation framework integrating GMMs with LLMs to address the challenges of class imbalance in unsupervised natural language processing tasks. Our approach targets underrepresented regions in the dataset, employing GMMs for precise cluster identification and LLMs for generating contextually relevant synthetic documents. Through extensive experiments on multiple imbalanced text datasets, we demonstrated that our method maintains clustering performance, improves the representation of minority classes, and enriches cluster interpretability. The results show that targeted data augmentation is effective in addressing imbalance without compromising the quality of clustering algorithms.

In our contribution, we relied on GMMs; it should be interesting to test other mixtures, such as von-Mises Fisher mixture models \cite{salah2017model,salah2017social,salah2019directional} or latent block models \cite{govaert2003clustering,nadif2005block,govaert2013co,hoseinipour2024sparse}. Additionally, a potential limitation of using LLM-generated data is the risk of reinforcing biases present in the training data of the LLM itself \cite{Shumailov2024}. While our approach ensures that generated text aligns with cluster themes, future work should explore ways to assess and mitigate biases in synthetic data.

\subsubsection*{Disclosure of Interests}
{\small
The authors have no competing interests to declare that are relevant to the content of this article.
}

\bibliographystyle{splncs04}
\bibliography{custom}

\end{document}